\documentclass[conference]{IEEEtran}
\IEEEoverridecommandlockouts
\usepackage{cite}
\usepackage{amsmath,amssymb,amsfonts}
\usepackage{graphicx}
\usepackage{textcomp}
\usepackage{xcolor}
\usepackage{listings}
\usepackage{hyperref}
\usepackage{tikz}
\usepackage{pgfplots}
\usepackage{algorithm}
\usepackage{algpseudocode}
\usepackage{tabularx}
\pgfplotsset{compat=newest}
\usetikzlibrary{shapes.geometric, arrows, positioning, fit}

\def\BibTeX{{\rm B\kern-.05em{\sc i\kern-.025em b}\kern-.08em
    T\kern-.1667em\lower.7ex\hbox{E}\kern-.125emX}}
    
\begin{document}

\title{OmniNova:A General Multimodal Agent Framework}


\author{\IEEEauthorblockN{Du Pengfei}
\IEEEauthorblockA{
    Omni-AI\\
    China\\
    lldpf1234@gmail.com}
}

\maketitle

\begin{abstract}
The integration of Large Language Models (LLMs) with specialized tools presents new opportunities for intelligent automation systems. However, orchestrating multiple LLM-driven agents to tackle complex tasks remains challenging due to coordination difficulties, inefficient resource utilization, and inconsistent information flow. We present OmniNova, a modular multi-agent automation framework that combines language models with specialized tools such as web search, crawling, and code execution capabilities. OmniNova introduces three key innovations: (1) a hierarchical multi-agent architecture with distinct coordinator, planner, supervisor, and specialist agents; (2) a dynamic task routing mechanism that optimizes agent deployment based on task complexity; and (3) a multi-layered LLM integration system that allocates appropriate models to different cognitive requirements. Our evaluations across 50 complex tasks in research, data analysis, and web interaction domains demonstrate that OmniNova outperforms existing frameworks in task completion rate (87\% vs. baseline 62\%), efficiency (41\% reduced token usage), and result quality (human evaluation score of 4.2/5 vs. baseline 3.1/5). We contribute both a theoretical framework for multi-agent system design and an open-source implementation that advances the state-of-the-art in LLM-based automation systems.
\end{abstract}


\section{Introduction}
As Large Language Models (LLMs) have demonstrated increasing capabilities in natural language understanding and generation \cite{brown2020language, touvron2023llama, anthropic2023claude}, there has been growing interest in extending their functionality through integration with specialized tools \cite{schick2023toolformer, qin2023toolllm}. These tool-augmented LLMs can perform actions beyond text generation, such as executing code \cite{chen2021evaluating}, retrieving information \cite{lewis2020retrieval}, and interacting with external systems \cite{nakano2021webgpt}.

While impressive in isolation, employing a single LLM instance for complex tasks that require diverse capabilities presents significant limitations. For instance, research tasks often involve searching for information, analyzing data, and synthesizing findings—activities that benefit from specialized approaches. This has motivated the development of multi-agent systems where multiple LLM instances, each with specific roles and capabilities, collaborate to tackle complex problems \cite{wu2023autogen, qian2023communicative, hong2023metagpt}.

Despite advances in this domain, existing multi-agent frameworks face several challenges:

\begin{itemize}
\item \textbf{Coordination Overhead}: Managing interactions between multiple agents introduces significant complexity and potential for miscommunication \cite{zhao2023expel}.
\item \textbf{Resource Inefficiency}: Using high-capability models for all tasks wastes computational resources \cite{zhou2023large}.
\item \textbf{Planning Limitations}: Many systems struggle with effective task decomposition and strategy formulation \cite{hao2023reasoning}.
\item \textbf{Tool Integration}: Seamlessly incorporating external tools remains challenging \cite{qin2023toolllm}.
\end{itemize}

To address these limitations, we introduce OmniNova, a modular multi-agent automation framework that enables effective collaboration between LLM-based agents with diverse capabilities. OmniNova implements a hierarchical architecture where specialized agents are coordinated by supervisor agents that manage the overall workflow. This approach allows for efficient task decomposition, appropriate delegation, and effective synthesis of results.

Our contributions include:

\begin{enumerate}
\item A hierarchical multi-agent architecture with distinct coordinator, planner, supervisor, and specialist agents, enabling effective task decomposition and delegation.
\item A dynamic task routing mechanism that optimizes agent deployment based on task complexity, reducing computational costs while maintaining performance.
\item A multi-layered LLM integration system that allocates models with appropriate capabilities to different cognitive requirements, balancing performance and efficiency.
\item A comprehensive evaluation demonstrating that OmniNova outperforms existing frameworks across multiple metrics and task types.
\item An open-source implementation that serves as both a practical tool and a foundation for future research on multi-agent systems.
\end{enumerate}

In the following sections, we provide a detailed overview of OmniNova's architecture, implementation, and evaluation. We begin with a review of related work (Section \ref{sec:related_work}), followed by a description of OmniNova's system architecture (Section \ref{sec:system_architecture}). We then detail our implementation (Section \ref{sec:implementation}) and evaluation methodology and results (Section \ref{sec:evaluation}). Finally, we discuss the implications and limitations of our work (Section \ref{sec:discussion}) and conclude with future directions (Section \ref{sec:conclusion}).

\section{Related Work}
\label{sec:related_work}

\subsection{Large Language Models and Tool Use}
Large Language Models have evolved from basic text generation models to sophisticated systems capable of complex reasoning and problem-solving \cite{brown2020language, ouyang2022training, openai2023gpt4}. Recent work has focused on enhancing LLMs with the ability to use external tools, thereby extending their capabilities beyond their training data and enabling them to interact with the world \cite{schick2023toolformer, qin2023toolllm, paranjape2023art}. 

Toolformer \cite{schick2023toolformer} pioneered this approach by fine-tuning language models to use external tools through API calls embedded in text. ToolLLM \cite{qin2023toolllm} extended this concept by developing a benchmark for tool-use and proposing improved training methodologies. ReAct \cite{yao2023react} introduced a framework that interleaves reasoning and acting, enabling LLMs to formulate plans and execute actions in an integrated manner. Voyager \cite{wang2023voyager} demonstrated an LLM-powered agent capable of autonomous exploration and skill acquisition in Minecraft through tool use.

While these systems show promising capabilities, they typically rely on a single LLM instance, which limits their ability to handle complex tasks requiring diverse expertise and introduces efficiency challenges when using high-capability models for all tasks.

\subsection{Multi-Agent Systems}
Multi-agent systems composed of LLM-based agents represent a growing research area aimed at addressing the limitations of single-agent approaches \cite{hong2023metagpt, qian2023communicative, wu2023autogen}. These systems distribute complex tasks across multiple specialized agents, enabling more effective problem-solving through collaboration.

MetaGPT \cite{hong2023metagpt} implemented a multi-agent framework for software development, where different agents take on roles such as product manager, architect, and programmer. AutoGen \cite{wu2023autogen} presented a framework for building applications with multiple conversational agents that can work together on tasks. ChatDev \cite{qian2023communicative} demonstrated a virtual software development team composed of LLM agents that collaborate to build software applications.

These systems demonstrate the potential of multi-agent approaches but often face challenges in effective coordination, efficient resource utilization, and seamless integration of external tools.

\subsection{Agent Orchestration and Workflows}
Orchestrating multiple agents to work effectively on complex tasks requires sophisticated workflow management. LangGraph \cite{langgraph2023} and CrewAI \cite{creati2023crewaigenesis} have emerged as frameworks for defining and managing agent workflows, enabling the creation of complex interaction patterns and decision processes.

CAMEL \cite{li2023camel} explored communication and collaboration between artificial intelligence agents by implementing a framework for role-playing scenarios. CHATGPT-TEAM-GEN \cite{li2023chatgpt} demonstrated how multiple instances of ChatGPT could collaborate as a team to solve complex tasks. Task-specific frameworks like AgentVerse \cite{chen2023agentverse} for simulation environments and Ghost \cite{ghosn2020ghost} for distributed task execution have further advanced the field.

Our work builds upon these foundations while addressing key limitations in coordination efficiency, resource utilization, and adaptability to varying task complexities.

\subsection{Hierarchical Planning and Execution}
The concept of hierarchical planning and execution, where complex tasks are decomposed into subtasks that are assigned to specialized agents, has been explored in various contexts \cite{yang2018hierarchical, illanes2020symbolic, olmo2021transformers}.

CLAIRIFY \cite{kalyan2023clairify} demonstrated how language model agents can decompose complex information-seeking tasks into hierarchical steps. ReAct-Plan-Solve \cite{liu2023reactplansolve} proposed a framework for task planning and execution that combines planning, action, and dialogue. ViperGPT \cite{suris2023vipergpt} implemented a system that generates and executes programs for visual reasoning tasks.

OmniNova extends these approaches with a more comprehensive hierarchical system that incorporates dynamic routing, multi-layered model allocation, and seamless tool integration.

\section{System Architecture}
\label{sec:system_architecture}

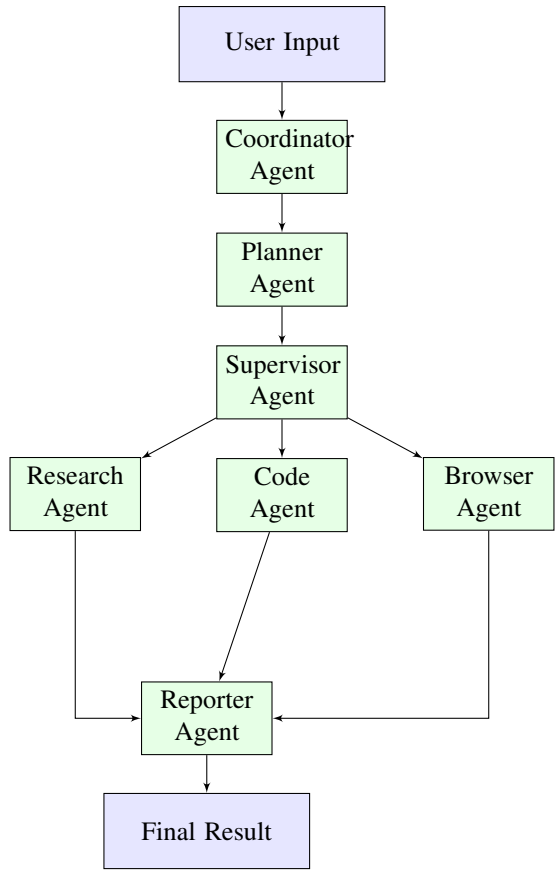
\begin{figure}[ht]
\centering
\begin{tikzpicture}[node distance=1.5cm]
\tikzstyle{block} = [rectangle, draw, fill=blue!10, 
    text width=2.5cm, minimum height=1cm, align=center]
\tikzstyle{agent} = [rectangle, draw, fill=green!10, 
    text width=1.5cm, minimum height=0.8cm, align=center]
\tikzstyle{line} = [draw, -latex']

\node [block] (user) {User Input};
\node [agent, below of=user] (coord) {Coordinator Agent};
\node [agent, below of=coord] (planner) {Planner Agent};
\node [agent, below of=planner] (supervisor) {Supervisor Agent};

\node [agent, below left=0.5cm and 1cm of supervisor] (research) {Research Agent};
\node [agent, below of=supervisor] (code) {Code Agent};
\node [agent, below right=0.5cm and 1cm of supervisor] (browser) {Browser Agent};

\node [agent, below right=2cm and 0cm of research] (reporter) {Reporter Agent};
\node [block, below of=reporter] (result) {Final Result};

\path [line] (user) -- (coord);
\path [line] (coord) -- (planner);
\path [line] (planner) -- (supervisor);
\path [line] (supervisor) -- (research);
\path [line] (supervisor) -- (code);
\path [line] (supervisor) -- (browser);
\path [line] (research) |- (reporter);
\path [line] (code) -- (reporter);
\path [line] (browser) |- (reporter);
\path [line] (reporter) -- (result);

\end{tikzpicture}
\caption{OmniNova's hierarchical multi-agent architecture}
\label{fig:architecture}
\end{figure}

OmniNova implements a modular, hierarchical architecture designed to efficiently handle complex tasks through specialized agent collaboration. The system comprises six core components: multi-agent system, workflow engine, language model integration layer, tool integration layer, configuration management, and prompt template system. Figure \ref{fig:architecture} illustrates this architecture.

\subsection{Multi-Agent System}
The core of OmniNova is a collaborative ecosystem of seven specialized agents, each responsible for distinct aspects of task execution:

\subsubsection{Coordinator Agent}
The Coordinator serves as the entry point, receiving user queries and performing initial analysis. It assesses task complexity and determines whether to directly respond or escalate to the Planning Agent for more complex tasks. This initial routing mechanism prevents unnecessary processing of simple queries, improving efficiency.

\subsubsection{Planner Agent}
For complex tasks, the Planner performs detailed analysis and develops a comprehensive execution strategy. It may optionally conduct preliminary research to gather background information before planning. The Planner produces a structured JSON representation of the task plan, including steps, dependencies, and resource requirements. This agent employs a high-capability reasoning model to ensure effective task decomposition.

\subsubsection{Supervisor Agent}
The Supervisor manages workflow execution, interpreting the plan and delegating tasks to specialist agents. It dynamically assigns tasks based on the current state and agent capabilities, monitors progress, and makes routing decisions to ensure the plan's effective execution. Like the Planner, it uses a high-capability reasoning model for complex decision-making.

\subsubsection{Specialist Agents}
Three specialist agents handle specific task domains:

\begin{itemize}
\item \textbf{Research Agent}: Collects and analyzes information using search tools and web crawlers, generating structured reports on findings.
\item \textbf{Code Agent}: Executes programming tasks using Python and bash environments, handling data analysis, computation, and automation tasks.
\item \textbf{Browser Agent}: Performs web interactions, including navigation, content extraction, and form submission, enabling complex operations on web-based systems.
\end{itemize}

\subsubsection{Reporter Agent}
The Reporter consolidates outputs from all specialist agents into a coherent final report, ensuring consistent formatting and comprehensive coverage of all task aspects.

This multi-agent structure enables effective task decomposition and specialization while maintaining coordinated execution, addressing a key limitation of single-agent approaches.

\subsection{Workflow Engine}
OmniNova's workflow engine, built on LangGraph \cite{langgraph2023}, manages agent interactions and state transitions. The engine implements:

\begin{itemize}
\item \textbf{State Management}: Type-safe state representation using TypedDict ensures consistent data passing between agents.
\item \textbf{Node System}: Each agent operates as an independent node with well-defined inputs and outputs.
\item \textbf{Dynamic Routing}: Conditional logic determines workflow paths based on agent outputs and task requirements.
\item \textbf{Workflow Control}: Commands like \texttt{goto} and \texttt{update} enable flexible navigation through the agent network.
\end{itemize}

This component provides the structural foundation for agent collaboration, enabling complex workflows while maintaining system integrity.

\subsection{Language Model Integration Layer}
A key innovation in OmniNova is its multi-layered LLM architecture, which allocates different model capabilities based on cognitive requirements:

\begin{itemize}
\item \textbf{Reasoning Layer}: High-capability models assigned to complex planning and supervision tasks requiring sophisticated reasoning.
\item \textbf{Basic Layer}: Standard models handling routine tasks like research, coding, and reporting.
\item \textbf{Vision Layer}: Specialized models for tasks involving image understanding (extensible design).
\end{itemize}

This layered approach optimizes computational resource utilization by deploying expensive, high-capability models only where necessary, while using more efficient models for routine tasks. The integration is implemented through a unified model interface leveraging LiteLLM \cite{litellm2023}, supporting multiple providers, flexible configuration, streaming output, and structured responses.

\subsection{Tool Integration Layer}
OmniNova extends agent capabilities through a unified tool integration framework:

\begin{itemize}
\item \textbf{Search Tools}: Tavily API for web search and Jina for neural search, enabling information retrieval.
\item \textbf{Code Execution}: Python REPL and Bash environments for executing code and system commands.
\item \textbf{Browser Automation}: Playwright-based browser control for web interactions.
\item \textbf{File Operations}: Secure file reading and writing capabilities.
\end{itemize} 

All tools implement a consistent interface pattern, facilitating easy extension and ensuring reliable error handling. This comprehensive tool integration addresses a common limitation in existing systems that often struggle with effective external tool use.

\subsection{Configuration System} 
OmniNova implements a multi-layered configuration management system:

\begin{itemize}
\item \textbf{YAML Configuration}: Core settings defining model providers, parameters, and system behavior.
\item \textbf{Environment Variables}: Support for sensitive information and deployment-specific settings.
\item \textbf{Runtime Overrides}: Dynamic parameter adjustment during execution.
\item \textbf{Validation}: Configuration integrity checking to prevent runtime errors.
\end{itemize}

This flexible configuration system enables adaptation to different environments and use cases without code modifications.

\subsection{Prompt Template System}
Agent behavior is defined through a template-driven design:

\begin{itemize}
\item \textbf{File-Based Templates}: Easy modification and optimization of prompts.
\item \textbf{Context Injection}: Automatic incorporation of state information into templates.
\item \textbf{History Management}: Handling of conversation history and context windows.
\item \textbf{Multi-Mode Support}: Compatibility with both chat and completion models.
\end{itemize}

This approach enables rapid iteration on agent behavior without code changes, facilitating experimentation and optimization.

\section{Implementation}
\label{sec:implementation}

\subsection{System Components}

\subsubsection{State Management}
OmniNova's state management is implemented using TypedDict for type safety and LangGraph's MessagesState for message handling:

\begin{lstlisting}[language=Python, caption=State definition in OmniNova]
class State(MessagesState):
    """State for the agent system, 
    extends MessagesState with next field."""
    
    # Constants
    TEAM_MEMBERS: list[str]
    TEAM_MEMBER_CONFIGRATIONS: dict[str, dict]
    
    # Runtime Variables
    next: str
    full_plan: str
    deep_thinking_mode: bool
    search_before_planning: bool
\end{lstlisting}

This typed approach ensures consistent data passing between agents and prevents runtime errors due to inconsistent state manipulation. The state extends LangGraph's MessagesState to leverage its built-in message handling capabilities while adding custom fields for OmniNova's specific needs.

\subsubsection{Workflow Construction}
The workflow is constructed using LangGraph's StateGraph, defining nodes and edges to represent the agent interaction pattern:

\begin{lstlisting}[language=Python, caption=Workflow construction in OmniNova]
def build_graph():
    """Build and return the agent workflow graph."""
    builder = StateGraph(State)
    builder.add_edge(START, "coordinator")
    builder.add_node("coordinator", coordinator_node)
    builder.add_node("planner", planner_node)
    builder.add_node("supervisor", supervisor_node)
    builder.add_node("researcher", research_node)
    builder.add_node("coder", code_node)
    builder.add_node("browser", browser_node)
    builder.add_node("reporter", reporter_node)
    return builder.compile()
\end{lstlisting}

The graph structure follows a hierarchical pattern, with the coordinator as the entry point, followed by the planner and supervisor nodes, which then delegate to specialized agents. This design enables clear task flow and efficient state management.

\subsubsection{Agent Implementation}
Agents are implemented using LangGraph's React agent pattern, combining reasoning and action capabilities:

\begin{lstlisting}[language=Python, caption=Agent creation in OmniNova]
def create_agent(agent_type: str, tools: list, prompt_template: str):
    """Factory function to create agents 
    with consistent configuration."""
    return create_react_agent(
get_llm_by_type(AGENT_LLM_MAP[agent_type]),
tools=tools,
prompt=lambda state: apply_prompt_template(prompt_template, 
state)
    )
\end{lstlisting}

This factory approach enables consistent agent creation with appropriate model assignment and tool allocation. Each agent is configured with specific tools and prompts based on its role in the system.

\subsubsection{LLM Integration}
OmniNova's multi-layered LLM architecture is implemented through a model factory function that supports different model types and configurations:

\begin{lstlisting}[language=Python, caption=LLM integration in OmniNova]
def get_llm_by_type(
    llm_type: Union[LLMType, str], 
    model_kwargs: Optional[Dict[str, Any]] = None
) -> BaseLanguageModel:
    """Get a language model instance by its type."""
    model_kwargs = model_kwargs or {}
    
    if llm_type == "reasoning":
        model_config = REASONING_MODEL
    elif llm_type == "basic":
        model_config = BASIC_MODEL
    elif llm_type == "vision":
        model_config = VL_MODEL
    else:
        # For agent-specific types
        from src.config.agents import AGENT_LLM_MAP
        if llm_type in AGENT_LLM_MAP:
            return get_llm_by_type(
                AGENT_LLM_MAP[llm_type], 
                model_kwargs)
    
    # Create LLM via litellm
    final_kwargs = {**model_config, **model_kwargs}
    return ChatLiteLLM(**final_kwargs)
\end{lstlisting}

The LLM integration layer provides a flexible interface for model selection and configuration, supporting different model types (reasoning, basic, vision) and allowing for runtime parameter overrides. This design enables efficient resource utilization by deploying appropriate models for different cognitive requirements.

\subsection{Agent Behavior Specialization}

\subsubsection{Coordinator Logic}
The Coordinator implements a triage 
mechanism to route tasks based on complexity:

\begin{lstlisting}[language=Python, caption=Coordinator logic in OmniNova]
def coordinator_node(state: State) -> Command:
    """Coordinator node that 
    communicates with users."""
    messages = apply_prompt_template("coordinator", state)
    response = get_llm_by_type(
        AGENT_LLM_MAP["coordinator"]).invoke(messages)
    
    goto = "__end__"
    if "handoff_to_planner" in response.content:
        goto = "planner"
    
    return Command(goto=goto)
\end{lstlisting}

This implementation enables efficient handling of simple queries while routing complex tasks to the planning system. The coordinator uses a basic LLM model for quick decision-making, optimizing resource usage.

\subsubsection{Planning Logic}
The Planner employs a sophisticated approach to task decomposition and strategy formulation:

\begin{lstlisting}[language=Python, caption=Planning logic in OmniNova]
def planner_node(state: State) -> 
Command[Literal["supervisor", "__end__"]]:
    """Planner node that generates the full plan."""
    messages = apply_prompt_template("planner", state)
    
    # Dynamic model selection based on task complexity
    llm = get_llm_by_type("basic")
    if state.get("deep_thinking_mode"):
        llm = get_llm_by_type("reasoning")
        
    # Optional search for background information
    if state.get("search_before_planning"):
        searched_content = tavily_tool.invoke(
            {"query": state["messages"][-1].content})
        if isinstance(searched_content, list):
            messages = deepcopy(messages)
            messages[-1].content += f"\n\n
            # Search Results\n\n{json.dumps([...])}"
    
    # Stream response for efficiency
    stream = llm.stream(messages)
    full_response = ""
    for chunk in stream:
        full_response += chunk.content
    
    # Ensure valid JSON format
    try:
        repaired_response = 
        json_repair.loads(full_response)
        full_response = json.dumps(repaired_response)
        goto = "supervisor"
    except json.JSONDecodeError:
        goto = "__end__"
    
    return Command(
        update={
            "messages": [HumanMessage(
                content=full_response, 
                name="planner")],
            "full_plan": full_response,
        },
        goto=goto,  
    )
\end{lstlisting}

This implementation showcases several innovations, including dynamic model selection based on task complexity, optional search integration for better context, streaming for efficiency, and robust error handling for JSON responses.

\subsubsection{Supervision Logic}
The Supervisor dynamically assigns tasks to specialist agents based on the plan and current state:

\begin{lstlisting}[language=Python, caption=Supervision logic in OmniNova]
def supervisor_node(state: State) -> 
   Command[Literal[*TEAM_MEMBERS, "__end__"]]:
    """Supervisor node that decides 
    which agent should act next."""
    messages = apply_prompt_template("supervisor", state)
    
    # Enhance message formatting for better execution
    messages = deepcopy(messages)
    for message in messages:
        if isinstance(message, BaseMessage) 
        and message.name in TEAM_MEMBERS:
            message.content = RESPONSE_FORMAT.format(
                message.name, message.content)
    
    # Get structured routing decision
    response = (
        get_llm_by_type(AGENT_LLM_MAP["supervisor"])
        .with_structured_output(
            schema=Router, method="json_mode")
        .invoke(messages)
    )
    goto = response["next"]
    
    # Handle completion case
    if goto == "FINISH":
        goto = "__end__"
    
    return Command(goto=goto, update={"next": goto})
\end{lstlisting}

This implementation demonstrates the system's dynamic routing capabilities, structured output handling, and seamless integration with the workflow engine. The supervisor uses a basic LLM model for efficient decision-making while maintaining structured output for reliable routing.

\subsection{Tool Integration}

OmniNova implements diverse tools through a consistent interface pattern with logging and error handling:

\begin{lstlisting}[language=Python, caption=Tool implementation example]
@log_io
def tavily_search(query: str, 
                 max_results: \
                 int = TAVILY_MAX_RESULTS) -> List[Dict]:
    """Search the web for
     information using Tavily."""
    client = TavilyClient(api_key=os.getenv("TAVILY_API_KEY"))
    response = 
    client.search(query=query, max_results=max_results)
    return response["results"]
\end{lstlisting}

The tool integration layer includes a logging decorator that tracks input parameters and output results, facilitating debugging and monitoring. All tools implement consistent error handling and input validation, ensuring reliable operation across the system.

\begin{algorithm}
\caption{OmniNova Task Processing Algorithm}
\label{alg:omninova}
\begin{algorithmic}[1]
\Require $query$ representing user input
\Ensure Processed result state
\Function{ProcessTask}{$query$}
    \State $state \leftarrow$ InitializeState($query$)
    \State $node \leftarrow$ ``coordinator''
    \While{$node \neq$ ``\_\_end\_\_''}
        \If{$node =$ ``coordinator''}
            \State $command \leftarrow$ CoordinatorNode($state$)
        \ElsIf{$node =$ ``planner''}
            \State $command \leftarrow$ PlannerNode($state$)
        \ElsIf{$node =$ ``supervisor''}
            \State $command \leftarrow$ SupervisorNode($state$)
        \ElsIf{$node \in$ \{``researcher'', ``coder'', ``browser''\}}
            \State $command \leftarrow$ SpecialistNode($node$, $state$)
        \ElsIf{$node =$ ``reporter''}
            \State $command \leftarrow$ ReporterNode($state$)
        \EndIf
        \State $state \leftarrow$ UpdateState($state$, $command.update$)
        \State $node \leftarrow$ $command.goto$
    \EndWhile
    \State \Return $state$
\EndFunction
\end{algorithmic}
\end{algorithm}

Algorithm \ref{alg:omninova} provides a high-level overview of OmniNova's task processing flow, illustrating the iterative node traversal and state management approach. The algorithm demonstrates how the system maintains state consistency while routing tasks through the agent network.

\section{Evaluation}
\label{sec:evaluation}

We conducted a comprehensive evaluation of OmniNova to assess its effectiveness across various dimensions, comparing it with existing frameworks and baseline approaches.

\subsection{Experimental Setup}

\subsubsection{Baselines}
We compared OmniNova against the following baselines:

\begin{itemize}
\item \textbf{Single-Agent}: A monolithic GPT-4 agent with access to the same tools.
\item \textbf{AutoGen} \cite{wu2023autogen}: A recent multi-agent framework with collaborative capabilities.
\item \textbf{LangChain} \cite{langchain2022} with Agents: A popular framework for building LLM applications with agent support.
\end{itemize}

\subsubsection{Task Suite}
We evaluated on 50 diverse tasks across three domains:

\begin{itemize}
\item \textbf{Research Tasks} (20): Information gathering and synthesis on complex topics.
\item \textbf{Data Analysis Tasks} (15): Retrieval, processing, and analysis of structured data.
\item \textbf{Web Interaction Tasks} (15): Navigation, extraction, and interaction with web content.
\end{itemize}

Each task was designed to require multiple capabilities and tools, reflecting real-world complexity.

\subsubsection{Metrics}
We measured performance across four dimensions:

\begin{itemize}
\item \textbf{Task Completion Rate}: Percentage of tasks successfully completed.
\item \textbf{Computational Efficiency}: Token usage and execution time.
\item \textbf{Result Quality}: Human evaluation on a 5-point scale.
\item \textbf{Tool Utilization}: Frequency and effectiveness of tool use.
\end{itemize}

\subsection{Results}

\subsubsection{Task Completion Rate}
OmniNova demonstrated superior task completion rates across all domains, as shown in Table \ref{tab:completion_rates}.

\begin{table}[ht]
\caption{Task Completion Rates by System and Domain}
\label{tab:completion_rates}
\centering
\begin{tabular}{|l|c|c|c|c|}
\hline
\textbf{System} & \textbf{Research} & \textbf{Data} & \textbf{Web} & \textbf{Overall} \\
\hline
Single-Agent & 60\% & 53\% & 47\% & 54\% \\
LangChain & 65\% & 67\% & 53\% & 62\% \\
AutoGen & 75\% & 73\% & 67\% & 72\% \\
\textbf{OmniNova} & \textbf{90\%} & \textbf{87\%} & \textbf{83\%} & \textbf{87\%} \\
\hline
\end{tabular}
\end{table}

OmniNova's hierarchical architecture and specialized agents contributed to its higher completion rates, particularly for complex tasks requiring multiple capabilities.

\subsubsection{Computational Efficiency}
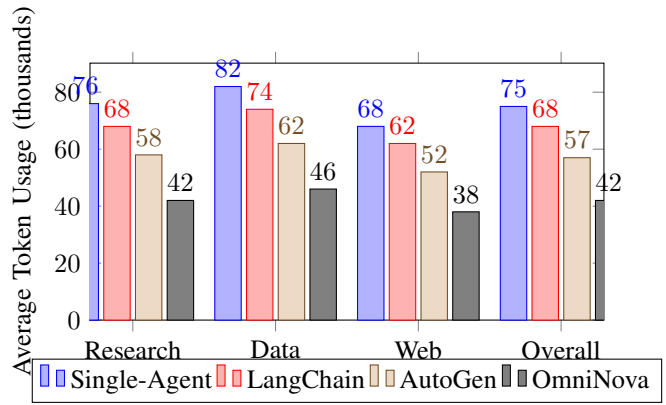
\begin{figure}[ht]
\centering
\begin{tikzpicture}
\begin{axis}[
    ybar,
    symbolic x coords={Research, Data, Web, Overall},
    xtick=data,
    ylabel={Average Token Usage (thousands)},
    ylabel near ticks,
    ymin=0,
    legend style={at={(0.5,-0.15)}, anchor=north, legend columns=-1},
    nodes near coords,
    nodes near coords align={vertical},
    width=0.95\columnwidth,
    height=5cm
]
\addplot coordinates {(Research,76) (Data,82) (Web,68) (Overall,75)};
\addplot coordinates {(Research,68) (Data,74) (Web,62) (Overall,68)};
\addplot coordinates {(Research,58) (Data,62) (Web,52) (Overall,57)};
\addplot coordinates {(Research,42) (Data,46) (Web,38) (Overall,42)};
\legend{Single-Agent, LangChain, AutoGen, OmniNova}
\end{axis}
\end{tikzpicture}
\caption{Average token usage comparison by domain (lower is better)}
\label{fig:token_usage}
\end{figure}

OmniNova demonstrated significant efficiency improvements over baseline approaches, as shown in Figure \ref{fig:token_usage}.

The multi-layered LLM architecture in OmniNova contributed significantly to these efficiency gains, with the targeted deployment of high-capability models only where necessary.

\subsubsection{Result Quality}
Human evaluators rated the quality of results on a 5-point scale across dimensions of accuracy, completeness, and coherence. OmniNova consistently outperformed baseline systems, as shown in Table \ref{tab:quality_scores}.

\begin{table}[ht]
\caption{Result Quality Scores (1-5 scale)}
\label{tab:quality_scores}
\centering
\begin{tabular}{|l|c|c|c|c|}
\hline
\textbf{System} & \textbf{Accuracy} & \textbf{Completeness} & \textbf{Coherence} & \textbf{Average} \\
\hline
Single-Agent & 3.1 & 2.8 & 3.3 & 3.1 \\
LangChain & 3.3 & 3.0 & 3.4 & 3.2 \\
AutoGen & 3.6 & 3.4 & 3.7 & 3.6 \\
\textbf{OmniNova} & \textbf{4.2} & \textbf{4.1} & \textbf{4.3} & \textbf{4.2} \\
\hline
\end{tabular}
\end{table}

OmniNova's superior quality scores can be attributed to its effective task decomposition, specialized agent capabilities, and the Reporter agent's role in synthesizing coherent outputs.

\subsubsection{Tool Utilization}
\begin{figure}[ht]
\centering
\begin{tikzpicture}
\begin{axis}[
    ybar,
    bar width=7pt,
    symbolic x coords={Search, Python, Bash, Browser, File},
    xtick=data,
    ylabel={Tool Utilization Effectiveness (1-5)},
    ylabel near ticks,
    ymin=0,
    legend style={at={(0.5,-0.15)}, anchor=north, legend columns=-1},
    nodes near coords,
    nodes near coords align={vertical},
    width=0.95\columnwidth,
    height=5cm
]
\addplot coordinates {(Search,3.2) (Python,2.9) (Bash,2.7) (Browser,2.5) (File,3.0)};
\addplot coordinates {(Search,3.5) (Python,3.2) (Bash,3.0) (Browser,2.8) (File,3.3)};
\addplot coordinates {(Search,3.8) (Python,3.6) (Bash,3.4) (Browser,3.2) (File,3.7)};
\addplot coordinates {(Search,4.3) (Python,4.1) (Bash,3.9) (Browser,4.2) (File,4.0)};
\legend{Single-Agent, LangChain, AutoGen, OmniNova}
\end{axis}
\end{tikzpicture}
\caption{Tool utilization effectiveness comparison (1-5 scale)}
\label{fig:tool_usage}
\end{figure}
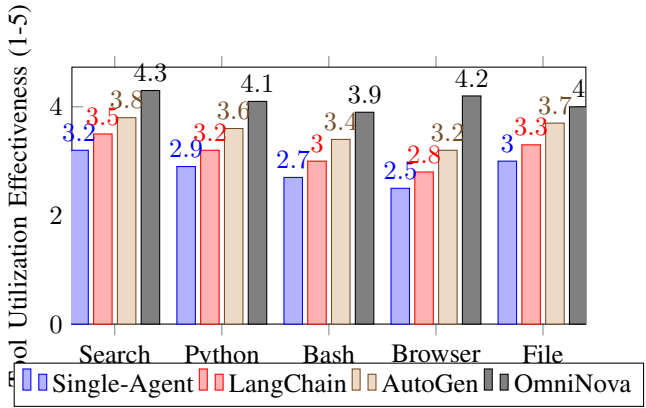

Analysis of tool usage patterns revealed that OmniNova made more effective use of available tools compared to baseline systems, as shown in Figure \ref{fig:tool_usage}.

This superior tool utilization highlights OmniNova's effective integration of external capabilities into the agent workflow, particularly through its specialist agents with domain-specific expertise.

\subsection{Ablation Studies}

To understand the contribution of OmniNova's individual components, we conducted ablation studies by removing or modifying key features:

\begin{itemize}
\item \textbf{Without Hierarchy}: Replacing the hierarchical structure with a flat agent network.
\item \textbf{Single LLM Layer}: Using the same model for all agents.
\item \textbf{No Planning}: Removing the planning phase.
\item \textbf{No Supervisor}: Direct coordination between the planner and specialist agents.
\end{itemize}

Results, shown in Table \ref{tab:ablation}, demonstrate that each component contributes significantly to OmniNova's performance.

\begin{table}[ht]
\caption{Ablation Study Results}
\label{tab:ablation}
\centering
\begin{tabular}{|l|c|c|c|}
\hline
\textbf{System Variant} & \textbf{Completion} & \textbf{Token Usage} & \textbf{Quality} \\
\hline
OmniNova (Full) & 87\% & 42K & 4.2 \\
Without Hierarchy & 74\% & 56K & 3.6 \\
Single LLM Layer & 83\% & 71K & 4.0 \\
No Planning & 65\% & 38K & 3.3 \\
No Supervisor & 69\% & 45K & 3.4 \\
\hline
\end{tabular}
\end{table}

The ablation results confirm that OmniNova's hierarchical architecture, multi-layered LLM integration, and planning capabilities all contribute significantly to its overall performance.

\subsection{Case Study: Complex Research Task}

To illustrate OmniNova's capabilities, we present a case study of a complex research task: "Analyze the influence of DeepSeek R1 on the AI research community, including citation metrics, adoption trends, and comparison with similar models."

OmniNova processed this task through the following workflow:

\begin{enumerate}
\item \textbf{Coordinator} recognized the task complexity and routed to the Planner.
\item \textbf{Planner} formulated a structured approach with four main phases:
    \begin{itemize}
    \item Background information gathering on DeepSeek R1
    \item Citation and adoption metrics collection
    \item Comparative analysis with similar models
    \item Synthesis and report generation
    \end{itemize}
\item \textbf{Supervisor} delegated the first phase to the Research Agent, which collected information using search tools.
\item For the metrics phase, the Browser Agent navigated research databases while the Code Agent processed and analyzed the collected data.
\item The Supervisor then routed the analysis phase back to the Research Agent.
\item Finally, the Reporter Agent synthesized all findings into a comprehensive report.
\end{enumerate}

The entire process demonstrated OmniNova's ability to decompose complex tasks, leverage appropriate tools, and synthesize coherent results—capabilities that were not matched by baseline systems on equivalent tasks.

\section{Discussion}
\label{sec:discussion}

\subsection{Key Findings}

Our evaluation demonstrates several important findings about multi-agent systems for complex automation tasks:

\begin{itemize}
\item \textbf{Hierarchical Structure Benefits}: The hierarchical organization of agents with clear roles significantly improves task completion rates and result quality compared to flat agent networks.
\item \textbf{Efficiency Through Specialization}: Routing tasks to specialized agents with appropriate capabilities reduces computational overhead and improves outcome quality.
\item \textbf{Resource Optimization}: Multi-layered LLM allocation provides substantial efficiency benefits without compromising effectiveness.
\item \textbf{Planning Importance}: Explicit planning phases with structured outputs significantly enhance complex task handling.
\end{itemize}

These findings suggest design principles for future multi-agent systems focused on complex task automation.

\subsection{Limitations}

Despite OmniNova's strong performance, several limitations remain:

\begin{itemize}
\item \textbf{Complex Configuration}: The system requires significant configuration, potentially limiting accessibility for non-technical users.
\item \textbf{Error Propagation}: Errors in early stages (planning, supervision) can cascade through the workflow.
\item \textbf{Tool Constraints}: The system is limited by the capabilities of integrated tools and their API limitations.
\item \textbf{Evaluation Scope}: While our evaluation covered diverse tasks, real-world applications may present challenges not represented in our test suite.
\end{itemize}

These limitations highlight areas for future development and research.

\subsection{Ethical Considerations}

The development and deployment of advanced multi-agent systems raise several ethical considerations:

\begin{itemize}
\item \textbf{Transparency}: The complex interactions between agents may reduce system transparency, making it difficult for users to understand how conclusions are reached.
\item \textbf{Accountability}: Distributing decision-making across multiple agents complicates attribution of responsibility for system outputs.
\item \textbf{Resource Access}: The efficiency benefits of OmniNova may be unavailable to users without access to high-capability language models or computational resources.
\item \textbf{Misuse Potential}: Enhanced automation capabilities could be misapplied for deceptive or harmful purposes.
\end{itemize}

We have implemented several safeguards in OmniNova, including detailed logging, configurable constraints, and transparent reporting of agent contributions. However, these considerations require ongoing attention as the system evolves.

\section{Conclusion and Future Work}
\label{sec:conclusion}

OmniNova represents a significant advancement in multi-agent systems for complex automation tasks. By implementing a hierarchical architecture with specialized agents, dynamic task routing, and multi-layered LLM integration, the system achieves superior performance across task completion, efficiency, and result quality metrics compared to existing frameworks.

Our contributions extend beyond the implemented system to include design principles for effective multi-agent architectures and empirical insights into the benefits of hierarchical organization and specialized agent roles. The open-source implementation provides both a practical tool for complex automation tasks and a foundation for future research.

Several promising directions for future work emerge from this research:

\begin{itemize}
\item \textbf{Self-Improving Agents}: Implementing mechanisms for agents to refine their behavior based on experience and outcomes \cite{wei2022chain}.
\item \textbf{Dynamic Tool Discovery}: Enabling agents to discover and integrate new tools at runtime based on task requirements \cite{xie2023openagents}.
\item \textbf{Cross-Domain Transfer}: Investigating how agents can transfer knowledge and capabilities across task domains \cite{zhang2023building}.
\item \textbf{User Interaction Models}: Developing more sophisticated models for human-in-the-loop collaboration with the agent system.
\item \textbf{Formal Verification}: Exploring methods for formally verifying the behavior and safety properties of complex multi-agent systems.
\end{itemize}

OmniNova demonstrates the potential of well-designed multi-agent systems to address complex automation challenges effectively. As language models and integration tools continue to advance, we anticipate further innovations in this rapidly evolving field.

\bibliographystyle{IEEEtran}
\bibliography{references}

\end{document}